\documentclass[10pt,twocolumn,letterpaper]{article}

\usepackage{cvpr}
\usepackage{times}
\usepackage{epsfig}
\usepackage{graphicx}
\usepackage{amsmath}
\usepackage{amssymb}
\usepackage{multirow}
\usepackage{booktabs}
\usepackage{cite}

\usepackage[breaklinks=true,bookmarks=false]{hyperref}

\cvprfinalcopy 

\def\cvprPaperID{****} 

\begin{document}

\title{DCDLearn: Multi-order Deep Cross-distance Learning for Vehicle Re-Identification}

\author{Rixing Zhu\\
{\tt\small rixingzhu@chd.edu.cn}
\and
Jianwu Fang\\
{\tt\small fangjianwu@chd.edu.cn}
\and
Hongke Xu\\
{\tt\small xuhongke@chd.edu.cn}
\and
Hongkai Yu\\
{\tt\small h.yu19@csuohio.edu}
\and
Jianru Xue\\
{\tt\small jrxue@mail.xjtu.edu.cn}
}

\maketitle

\begin{abstract}
   Vehicle re-identification (Re-ID) has become a popular research topic owing to its practicability in intelligent transportation systems. Vehicle Re-ID suffers the numerous challenges caused by drastic variation in illumination, occlusions, background, resolutions, viewing angles, and so on. To address it, this paper formulates a multi-order deep cross-distance learning (\textbf{DCDLearn}) model for vehicle re-identification, where an efficient one-view CycleGAN model is developed to alleviate exhaustive and enumerative cross-camera matching problem in previous works and smooth the domain discrepancy of cross cameras. Specially, we treat the transferred images and the reconstructed images generated by one-view CycleGAN as multi-order augmented data for deep cross-distance learning, where the cross distances of multi-order image set with distinct identities are learned by optimizing an objective function with multi-order augmented triplet loss and center loss to achieve the camera-invariance and identity-consistency. Extensive experiments on three vehicle Re-ID datasets demonstrate that the proposed method achieves significant improvement over the state-of-the-arts, especially for the small scale dataset.
\end{abstract}

\section{Introduction}\label{section1}
Vehicle re-identification (Re-ID) can be treated as a cross-camera vehicle retrieval task, i.e., searching for the relevant images of a query vehicle from vehicle gallery, which is of great significance to traffic safety and management \cite{yanTWZH2017}. For vehicle identification, although license plate is a natural and unique information, and license plate recognition has already been widely utilized in transport operation systems \cite{Christos2008,WenLYZDS2011,GouWYL2016}, unfortunately, in many scenes, the license plate information is infeasible because of the various factors, such as low-resolution, occlusion, dim illumination and motion blur \cite{ZhouLS2018}, and even being removed or faked sometimes. Therefore, vehicle appearance instead is attracting increasing attention for vehicle Re-ID.

In vehicle Re-ID task, vehicle images are captured from non-overlapping camera views. Therefore, visual appearance of a vehicle often undergoes drastic variations in illumination, occlusions, background, and viewing angles, etc \cite{ShenXLYW2017}. In order to handle these challenges, existing approaches are typically struggling to overcome the variations by designing discriminative feature representation \cite{LiuLMF2016,ZhouS2018} or robust matching models. However, vehicle Re-ID faces a severe problem that each vehicle has only one or few shots in the gallery set, which is a mini-sample mining problem intrinsically. To address this problem, the methods based on generative adversarial network (GAN) \cite{goodfellow2014} are recently employed to perform domain adaptation or image-to-image style transfer for a data augmentation in Re-ID task \cite{ZhongZZLY2018,chungD2019}, which is promising in increasing data diversity to alleviate over-fitting. Among them, CycleGAN \cite{ZhuPIE2017} is a typical framework which generates a fake image of the real one and then reconstructs (generates) the real one reversely by the fake image with a circle consistency constraint. Here, we denote the fake image as $1^{st}$-order generated sample, the reconstructed one as $2^{nd}$-order generated sample, and real one as $0^{th}$-order sample. In the state-of-the-art Re-ID works \cite{DengZYKYJ2018,ZhongZZLY2019,chungD2019} inspired by CycleGAN, they only treat the $1^{st}$-order images as augmented data for model learning, and ignore the utilization of $2^{nd}$-order ones and the relationship between multi-order samples.

A single CycleGAN can learn one-to-one style transfer for one camera pair. In multi-shot Re-ID problem, each identity is captured by multiple cameras, which consequently needs multiple CycleGAN models to build an complete camera style-transferred network exploited by \cite{ZhongZZLY2019}. Undoubtedly, the task complexity and computing overhead will be significantly increased as the number of camera pairs. Moreover, it requires enough image pairs of corresponding cameras to facilitate the estimation of model parameters. However, it is prohibitive to collect well-annotated image datasets. Aiming at overcoming these limitations, we develop an efficient one-view CycleGAN to divide the training set into two parts randomly, and only use one CycleGAN model for cross-camera transfer in vehicle Re-ID task, where an identity constraint is introduced to preserve identity similarity in transferring, so the transferred images can keep the same identity with real ones.

In order to fully leverage the diversity of the augmented data by our one-view CycleGAN model, we propose a multi-order deep cross-distance learning (\textbf{DCDLearn}) model for vehicle re-identification, where $0^{th}$-order, $1^{st}$-order and $2^{nd}$-order images construct the cross distance set adaptively selected in model learning, and the $2^{nd}$-order images after two iterations of generation actually carry the information from both two separate domains (formed by two or multiple camera views) to be matched. In this way, it will improve the model robustness to adapt to the cross-camera discrepancy in vehicle Re-ID task. Furthermore, we also introduce a multi-order augmented center loss inspired by \cite{wen2016discriminative} to compact intra-variation of the multi-order images with the same identity. In the training process, DCDLearn model can learn the optimal vehicle representation by automatical cross-distance selection, which is useful for identifying hard pair of samples. Extensive experiments on three vehicle Re-ID datasets show a state-of-the-art performance of the proposed method.

In summary, the main \textbf{contributions} of this paper are:
\begin{itemize}
\item In order to avoid the complex multiple cameras matching problem, we design an efficient one-view CycleGAN strategy to generate the style-transferred images for multi-camera Re-ID, where an identity constraint is introduced to preserve vehicle identities in transferring and can smooth the disparity among different cameras.
\item We utilize both style-transferred and reconstructed images generated by one-view CycleGAN as multi-order augmented labeled data, and formulate a deep cross-distance learning model involving a multi-order augmented triplet loss and center loss function to learn the optimal vehicle representation of multi-order images, reducing the cross-camera discrepancy for vehicle Re-ID, which can give a reliable learning for hard samples.
\item Entensive experiments on three vehicle Re-ID datasets demonstrate state-of-the-art performance of the proposed method.
\end{itemize}
The rest of the paper is organized as follows. Section \ref{section2} presents the related works. A detailed description of the proposed method is presented in Section \ref{section3}. Section \ref{section4} gives the experimental results, and the conclusions of this work is presented in Section \ref{section5}.
\section{Related Work}\label{section2}
This work is closely related with vehicle Re-ID and image-to-image style transfer by GANs, as briefly discussed in the following subsections.
\subsection{Feature Learning in Vehicle Re-ID}
Due to drastic appearance variations of a vehicle, it is crucial to design robust and discriminative feature representation for vehicle Re-ID. Previous works \cite{shanSK2008, zapletalH2016, LiuLMF2016 } utilized color, texture and vehicle type to distinguish vehicle identities. With the rapid development of convolutional neural network, deep learning based methods for feature representation have shown prominent advantage over traditional hard-crafted features \cite{LiuLMM2016, LiuLMM2018}. For example, the works \cite{liuWPH2016,BaiLGWWD2018} designed triplet network to measure the VGG feature similarity of positive pairs and negative pairs, which takes intra-class variance and inter-class similarity into account. Specially, some methods also fuse deep features with traditional features for feature representation. For instances, Liu \textit{et al.} \cite{LiuLMF2016} extracted GoogLeNet deep features, color and SIFT features, and applied the late fusion strategy to calculate similarity scores for vehicle Re-ID. Tang \textit{et al.} \cite{TangWJZL2017} designed a multi-modal feature architecture, which integrated LBP texture map and Bag-of-Word-based \emph{Color Name} feature into an end-to-end convolutional neural network.

It is worthy noting that there are many unique patterns on the vehicle, such as vehicle logo, light and stickers. Therefore, local features can also be valuable for Re-ID task \cite{zhaoSWC2019,PengWZF2019}. For example, Li \textit{et al.} \cite{LiYJL2017} extracted local features from windscreen area and global vehicle type features to identify vehicles. He \textit{et al.} \cite{heLZT2019} proposed part-regularized discriminative feature preserving method to enhance the perceptive capability of subtle distinction.

In addition, the challenge due to large viewing angle change induces drastic intra-discrepancy in vehicle appearance. Therefore, robust feature representation for multiple views begins to draw large attention. Liu \textit{et al.} \cite{ZhouS2018} proposed a viewpoint-aware attentive multi-view inference (VAMI) model and an adversarial architecture to conduct multi-view feature inference. Zhu \textit{et al.} \cite{zhuZHLLCZ2019} designed quadruple directional deep learning networks to obtain quadruple directional deep learning features of vehicle images. Furthermore, Wang \textit{et al.} \cite{WangTLYY2017} exploited vehicle viewpoint attribute and proposed orientation invariant feature embedding module by using 20 vehicle key points.
\begin{figure*}
  \centering
  \includegraphics[width=\hsize]{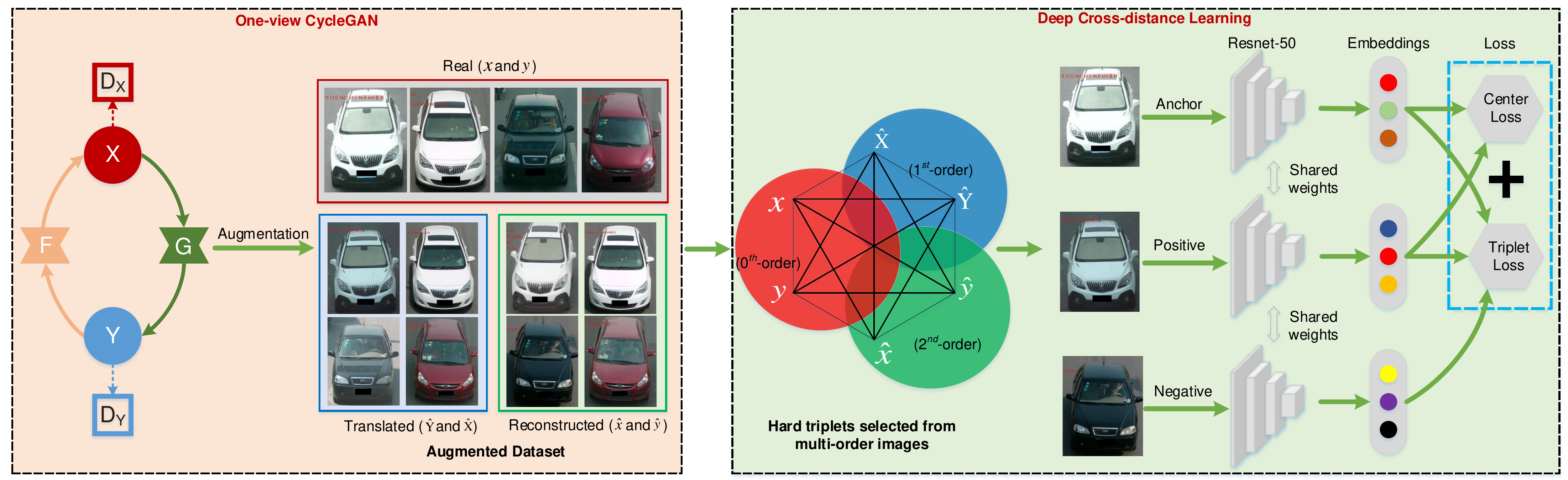}
  \caption{Overview of our pipeline. For the vehicle Re-ID task, we first train an efficient one-view CycleGAN to generate the transferred ($1^{st}$-order) and the reconstructed ($2^{nd}$-order) images from real ones ($0^{th}$-order). Then we feed these multi-order images into a deep cross-distance learning model involving a multi-order augmented triplet loss and center loss function to learn the optimal vehicle representations of multi-order images, where the cross distances of multi-order images are automatically selected in learning.}
  \label{fig1}
\end{figure*}

\subsection{Image-to-Image Style Transfer by GANs}
Recently, GANs \cite{goodfellow2014} have achieved outstanding success, and its many variants \cite{liW2016, ledig2017, yiZTG2017, kim2017, TaigmanPW2017} have been applied to image-to-image style transfer, cross-domain transfer, sketch-to-image generation, \emph{etc}. Especially, image-to-image style transfer has drawn much attention. Isola \textit{et al.} \cite{isola2017 } proposed a conditional GANs to learn a mapping from input to output images, which requires paired images in training process. To address complex pairing problem, Zhu \textit{et al.} \cite{ZhuPIE2017} proposed CycleGAN using cycle consistency loss to transfer images between two different domains with unpaired samples. Cross-domain style transfer methods transfer the domain (or style) of input image to another while maintaining the essential image content, which can also be regarded as image-to-image style transfer. Bousmalis \textit{et al.} \cite{bousmalis2017} proposed an unsupervised PixelDA model that transfers images of source domain to analog images in target domain. Choi \textit{et al.} \cite{ChoiCKHKC2018} designed a novel StarGAN that can perform image-to-image style transfer for multiple domains within only a single model.

Image-to-image style transfer methods have been introduced into Re-ID task \cite{WeiZGT2018, WangYYBS2018, tangZL2019, LiuZCHW2019}, which can reduce the risk of overfitting and alleviate the disparities among different cameras (or domains). Zhong \textit{et al.} \cite{ZhongZZLY2019} introduced CamStyle to perform multiple view learning and unsupervised domain adaptation. Deng \textit{et al.} \cite{DengZYKYJ2018} combined Siamese network and CycleGAN to preserve self-similarity and domain-dissimilarity in source domain and target domain. Chung \textit{et al.} \cite{chungD2019} proposed a similarity preserving StarGAN for multi-domain image-to-image style transfer. Zhou \textit{et al.} \cite {ZhouS2018} proposed a conditional feature-level generative network to transform single-view features into multi-view features. Our work aims to find the relationship of multi-order cross transferred images to promote the domain-similarity learning for vehicle Re-ID.

\section{The Proposed Method}\label{section3}
\subsection{Problem Formulation}
Given query vehicles, vehicle Re-ID aims to search the one(s) with the same identity (named as \emph{positive samples}) from the gallery set collected from other camera views. Commonly, the querying of each vehicle needs to find the positive samples from more massive negative ones with different identities. In order to obtain a preferable vehicle Re-ID model, we formulate the following objective function:
\begin{equation}
\begin{array}{l}
\mathop {{\rm{arg}}\min }\limits_{\boldsymbol{w}} \sum {\left[{\max {\kern 4pt}{d_p}({f_{\boldsymbol{w}}}(X^a;{\boldsymbol{w}}),{f_{\boldsymbol{w}}}(Y^{pos};{\boldsymbol{w}}))} \right.} \\
{\kern 38pt} \left. { - \min {\kern 4pt}  {d_n}({f_{\boldsymbol{w}}}(X^a;{\boldsymbol{w}}),{f_{\boldsymbol{w}}}(Y^{n{\rm{e}}g};{\boldsymbol{w}}))} \right],
\end{array}
\label{eq:1}
\end{equation}
where ${f_{\boldsymbol{w}}}(\cdot;{\boldsymbol{w}})$ specifies a feature embedding operation, and ${\boldsymbol{w}}$ is the parameters to be optimized. $X^a$, $Y^{pos}$ and $Y^{neg}$ are the query vehicle image, positive sample and negative sample, respectively. ${d_p}(\cdot,\cdot)$ and ${d_n}(\cdot,\cdot)$ represent the distance functions of vehicle identities. This equation is actually a triplet formulation and means that we want the maximum distance between the query vehicle image with positive samples to be smaller than the minimum distance with negative ones, and prefer a large margin. In this way, we can distinguish the hard vehicle images.

In Eq. \ref{eq:1}, it needs to put a triplet $\{X^a$, $Y^{pos}$,$Y^{neg}\}$ into each step of training, which faces another problem that it has limited positive identities, which may cause overfitting easily. An strategy is to enhance the diversity of identities. In this work, we design an efficient one-view CycleGAN for this purpose, and each vehicle image will generate a transferred image (one iteration from the real one) and a reconstructed image (two iterations from the real one), producing the so-called \emph{multi-order images}. Then, we further explore the cross-relation of the multi-order images to learn the optimal vehicle representation and reduce the cross-camera discrepancy for vehicle Re-ID. Fig. \ref{fig1} demonstrates the pipeline of the proposed method, and we will describe each module in following.
\begin{figure}
  \centering
  \includegraphics[width=0.85\hsize]{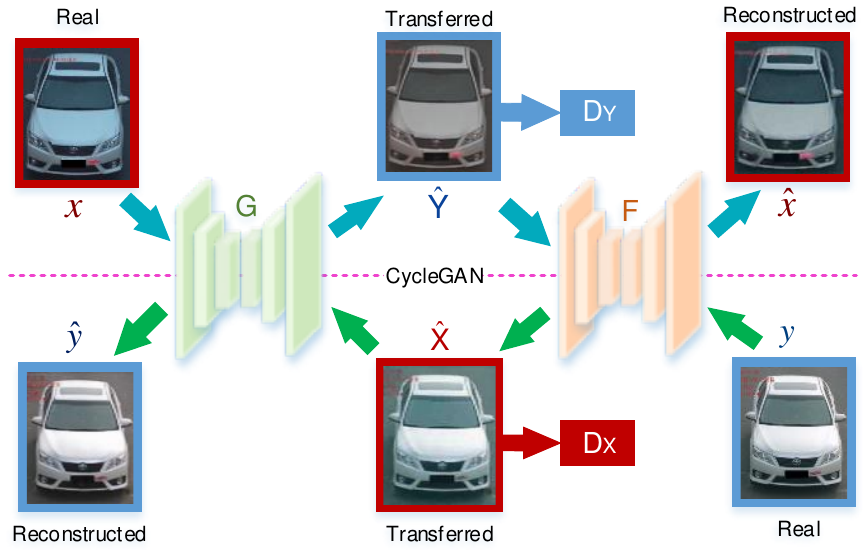}
  \caption{CycleGAN learns mapping functions $G$ and $F$ between domains $X$ and $Y$. $D_Y$ drives $G$ to transfer the style of vehicle images from $X$ to the outputs indistinguishable from domain $Y$, and vice versa for $D_Y$ and $F$.}
  \label{fig2}
\end{figure}
\subsection{One-view CycleGAN for Multiple Cameras}
For vehicle Re-ID task, CycleGAN aims to learn mapping functions between two domains $X$ and $Y$, where training samples $\left\{x_{i}\right\}_{i=1}^{N} \in X$ and $\left\{y_{j}\right\}_{j=1}^{M} \in Y$. CycleGAN is constructed by two generator-discriminator pairs $\{G, D_X\}$ and $\{F, D_Y\}$. The two generators $G$: $X$$\rightarrow$$Y$ and $F$: $Y$$\rightarrow$$X$ intend to generate similar images to the images from the other domain, while the two adversarial discriminators $D_X$ and $D_Y$ are utilized to differentiate whether images are transferred from the other domain. The overview of CylceGAN is illustrated in Fig.~\ref{fig2}.

In multi-camera Re-ID task, each vehicle is captured by at least two cameras, which needs to train multiple CycleGAN models to build an entire camera style-transferred network. For example, ${\rm C }_{20}^2 = 190$ different CycleGAN models may be needed for the VeRi dataset \cite{LiuLMM2016} with $20$ different cameras with prohibitive complexity. Although StarGAN \cite{ChoiCKHKC2018} developed a single model to perform image-to-image style transfer for multiple cameras, it will fail to estimate the enormous model parameters if the number of images from certain cameras is limited. In addition, collecting the camera label of each vehicle is laborious, such as VehicleID dataset which lacks camera label data. Therefore, there is no enough labeled pair of images from two cameras to achieve a practicable CycleGAN model.

In order to address these problem, we develop an efficient one-view CycleGAN for transferring the style of multiple cameras. Specifically, we divide the whole training set into two parts randomly. One is regarded as domain $X$, the other is regarded as domain $Y$. Although there are overlapped cameras between domain $X$ and $Y$, we attach an identity constraint to the loss function of CycleGAN for preserving the identities of vehicle images in different cameras, i.e., the transferred image of same vehicle are regarded as having the same ID whichever cameras they come from. The objective function of one-view CycleGAN is:

\begin{figure}
  \centering
  \includegraphics[width=0.85\hsize]{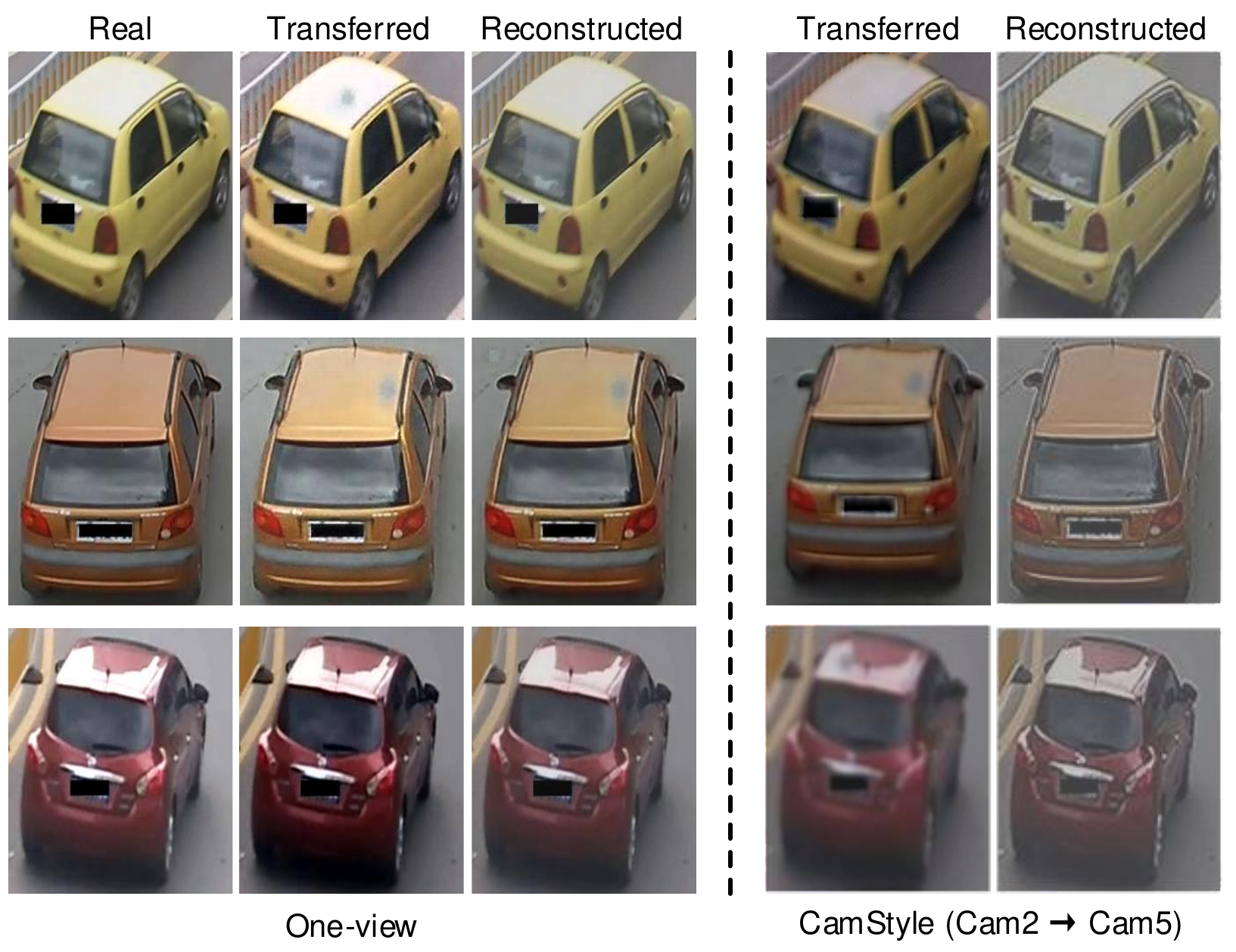}
  \caption{Examples of the transferred and reconstructed images after one-view CycleGAN (left) and original CycleGAN (right).}
  \label{fig3}
\end{figure}

\begin{equation}
\mathcal{\tilde L}\left(G, F, D_{X}, D_{Y}\right)=\mathcal{L}_{Gadv}+\mathcal{L}_{Fadv}+\alpha \mathcal{L}_{cyc}+\beta \mathcal{L}_{id}, \label{eq:2}
\end{equation}
where $\mathcal{L}_{id}$, $\mathcal{L}_{Gadv}$, $\mathcal{L}_{Fadv}$, and  $\mathcal{L}_{cyc}$ are the identity consistent loss between cameras, adversarial losses for generator-discriminator pairs $\{G, D_X\}$, $\{F, D_Y\}$, and the cycle consistent loss, respectively defined as:
\begin{equation}\small
\begin{aligned}
& \mathcal{L}_{id}= \mathbb{E}_{x \sim p_{x}} [||F(x)-x||_{1}] + {\kern 1pt} \mathbb{E}_{y \sim p_{y}} [||G(y)-y||_{1}] \\
& \mathcal{L}_{G a d v} = \mathbb{E}_{y \sim p_{y}}\left[\left(D_{Y}(y)-1\right)^{2}\right]+\mathbb{E}_{x \sim p_{x}}\left[\left(D_{Y}(G(x))^{2}\right]\right.\\
& \mathcal{L}_{F a d v} = \mathbb{E}_{x \sim p_{x}}\left[\left(D_{X}(x)-1\right)^{2}\right]+\mathbb{E}_{y \sim p_{y}}\left[\left(D_{X}(F(y))^{2}\right]\right.\\
& \mathcal{L}_{c y c}= \mathbb{E}_{x \sim p_{x}}[||F(G(x))-x||_{1}] +\mathbb{E}_{y \sim p_{y}}[||G(F(y))-y {\kern 1pt} ||_{1}] \\
\end{aligned}
\label{eq:3}
\end{equation}
$\alpha$ and $\beta$ are the trade-off parameters for controlling the relative importance of four losses.

Fig.~\ref{fig3} demonstrates a comparison of some examples of transferred and reconstructed images after one-view CycleGAN and original CycleGAN. From this figure, the qualitative superiority is validated.

\subsection{Deep Cross-distance Learning Re-ID Model}
\label{subsection:3}
In this paper, we employ one-view CycleGAN to serve as data augmentation. The new augmented dataset is a combination of the original images, the transferred images, and the reconstructed images. Since each augmented image retains the content of its original image, they are regarded as the same identity. In this manner, we leverage the new augmented dataset to optimizing Eq. \ref{eq:1}, and a triplet loss is a natural choice. Differently, we involve a multi-order augmented triplet loss penalty. What's more, we want the multi-order images with the same identity have a compact representation apart from other ones. A multi-order augmented center loss inspired by \cite{wen2016discriminative} is introduced and jointly optimized. Therefore, the full objective function is:
\begin{equation}\small
\begin{array}{l}
{{\cal L}_{t+c}}({\boldsymbol{w}}) = \sum\limits_{i,j=1}^K { {\left[ {} \right.m + \overbrace {\mathop {\max }\limits_i d\left( {f_{\boldsymbol{w}}\left( {x_i^a} \right),f_{\boldsymbol{w}}\left( {y_i^{pos}} \right)} \right)}^{\text{hardest{\kern 2pt} positive}}} } \\
{\kern 76pt}  - \underbrace {\mathop {\min }\limits_{i \neq j} d\left( {f_{\boldsymbol{w}} \left( {x_i^a} \right),f_{\boldsymbol{w}} \left( {y_j^{neg}} \right)} \right)}_{\text{hardest{\kern 2pt} negative}}] \\
{\kern 48pt} + \frac{\lambda}{2} \sum\limits_{i=1}^K ||f_{\boldsymbol{w}}\left( {x_i^a} \right)-\mathbf{c}_i||^2,
\end{array}
\label{eq:4}
\end{equation}
where $f_{\boldsymbol{w}}$($\cdot$) is the function to learn CNN embeddings of vehicle images, $d$($\cdot$,$\cdot$) is the cross-distance function between multi-order images, $m$ is a soft-margin threshold, $K$ denotes the number of identities in each batch in training. \{$x_i^a, y_i^{pos}, y_j^{neg}$\} denotes a triplet selected from the multi-order images in each batch, \{$x_i^a, y_i^{pos}$\} denotes a positive pair of images with same vehicle identity, while \{$x_i^a, y_j^{neg}$\} denotes a negative pair with different vehicle identities. The center $\mathbf{c}_i$ is computed by averaging the features of muti-order images with the same identity within a batch. The scalar $\lambda$ is used for balancing the multi-order augmented triplet loss and the multi-order augmented center loss. Notably, Eq. \ref{eq:4} picks the hardest triplet in each batch for each training step, and the hard triplet samples are picked among the real images ($0^{th}$-order), the transferred images ($1^{st}$-order), and the reconstructed images ($2^{nd}$-order) randomly selected in each batch. The hardness are computed by the largest Euclidean distance between embeddings of positive pairs or the smallest one between negative pairs.

\begin{figure}
  \centering
  \includegraphics[width=\hsize]{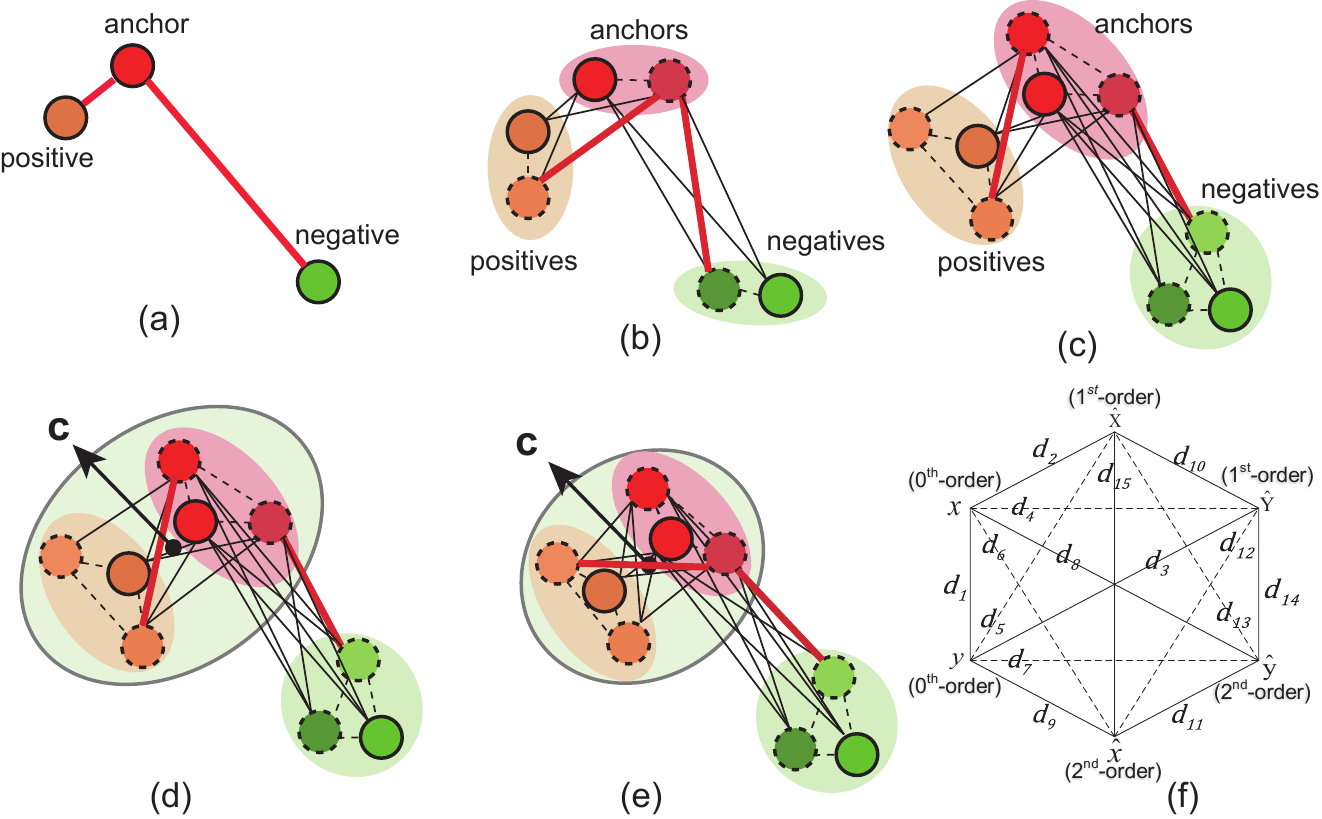}
  \caption{Illustration of multi-order augmented losses. Physical meaning of (a) original triplet loss, (b) one-order augmented triplet loss, (c) multi-order augmented triplet loss, (d) multi-order augmented triplet loss showing the representation center of the same identity, and (e) multi-order augmented triplet loss and center loss. (f) presents a Hexagram structure of cross distances of the embeddings of multi-order images. The dashed lines denote the relations between the samples having the same source of original image while the solid lines represent the relations between the samples coming from different domains. The largest intra-distance (LID) and smallest inter-distance (SID) are marked by red solid lines.}
    \vspace{-1em}
  \label{fig4}
\end{figure}

\textbf{Multi-order augmented triplet loss and center loss.} In Eq. \ref{eq:4}, there are two kinds of losses, i.e., the multi-order triplet loss and multi-order center loss. In order to illustrate them clear, we visualize their physical meanings in Fig. \ref{fig4}. We can see that multi-order vehicle images can 1) provide more diverse samples for training, and 2) more possibilities for harder triplet selection determined by the largest intra-distance (LID) and the smallest inter-distance (SID) than ever before, where multi-order images can give larger LID and smaller SID (Fig. \ref{fig4}(c)) than the ones in original triplet (Fig. \ref{fig4}(a)). As for the multi-order augmented center loss, it can compact the image representation to the center of the same identity, but might weaken the hardness of the triplet, shown by the shorten LID (Fig. \ref{fig4}(e)). Therefore, this work makes a trade-off for balancing the multi-order augmented triplet loss and multi-order augmented center loss.

\textbf{Interpretation of the cross distances.} To make the cross-distance between multi-order images clear, we re-phrase Fig. \ref{fig2} as Fig. \ref{fig4}(f) with a Hexagram structure. For a pair of domains, the similarity from $x\leftrightarrow y$ equals to $x/{\rm{\hat Y}}/\hat x \leftrightarrow y/{\rm{\hat X}}/\hat y$. According to the identity numbers $K$ in each batch, for the selection of \{$x_i^a, y_i^{pos}, y_j^{neg}$\}, we have $C_6^2$ positive pairs of \{$x_i^a, y_i^{pos}$\} and $C_{(K-1)\times 6}^1$ negative pairs of \{$x_i^a, y_j^{neg}$\}. Therefore, we largely boost the diversity of samples. Among the cross distances, $d_1,d_{10},d_{11}$ specify the intra-order cross distances within $0^{th}$-order, $1^{st}$-order and $2^{nd}$-order images, $d_2,d_3$ are the inter-order cross distances over $0^{th}$-order and $1^{st}$-order samples, $d_8,d_9$ denote the inter-order cross distances over $0^{th}$-order and $2^{nd}$-order images, and $d_{14},d_{15}$ denote the inter-order cross distances over $1^{st}$-order and $2^{nd}$-order images, respectively. Note that, $({\rm{\hat Y}},{\rm{\hat X}})$ and $(\hat x,\hat y)$ can be treated as $1^{st}$-order and $2^{nd}$-order \textbf{mirror} images in the reverse domain of $(x,y)$, respectively. Therefore, $d_{10}$ and $d_{11}$, to some extent, indicate the relations of samples within a common space approached from two different domains. Notably, as shown in Fig. \ref{fig4}(f), the relations marked by dashed lines cannot be considered in testing because the linked samples come from the same source.

\textbf{Training scheme.} In the training process, the proposed method consists of two phases. In the first phase, we adopt the one-view CycleGAN to generate augmented data, where the generator is comprised of $9$ residual blocks and four convolutions, while the discriminator is comprised of four convolutions and one fully connected layer, with the same configuration of CycleGAN \cite{ZhuPIE2017}. The generator and discriminator are trained by turns to optimize:
\begin{equation}
G^{*}, F^{*}=\arg \max _{D_{X}, D_{Y}} \min _{G, F} \mathcal{\tilde L}\left(G, F, D_{X}, D_{Y}\right).
\end{equation}

In the second phase, we adopt ResNet-50 \cite{heKaiming2016} as backbone for the embedding extraction of vehicle image, and then ImageNet pre-trained weights for triplet embedding learning are introduced to optimize:
\begin{equation}
{\boldsymbol{w}}^* = \arg \min {\cal L}_{t+c}({\boldsymbol{w}}).
\end{equation}
It can automatically learn the optimal vehicle representation with the deep cross-distance learning model.
\section{Experiments}\label{section4}
\subsection{Datasets and Evaluation Metrics}
\begin{figure}
  \centering
  \includegraphics[width=0.9\hsize]{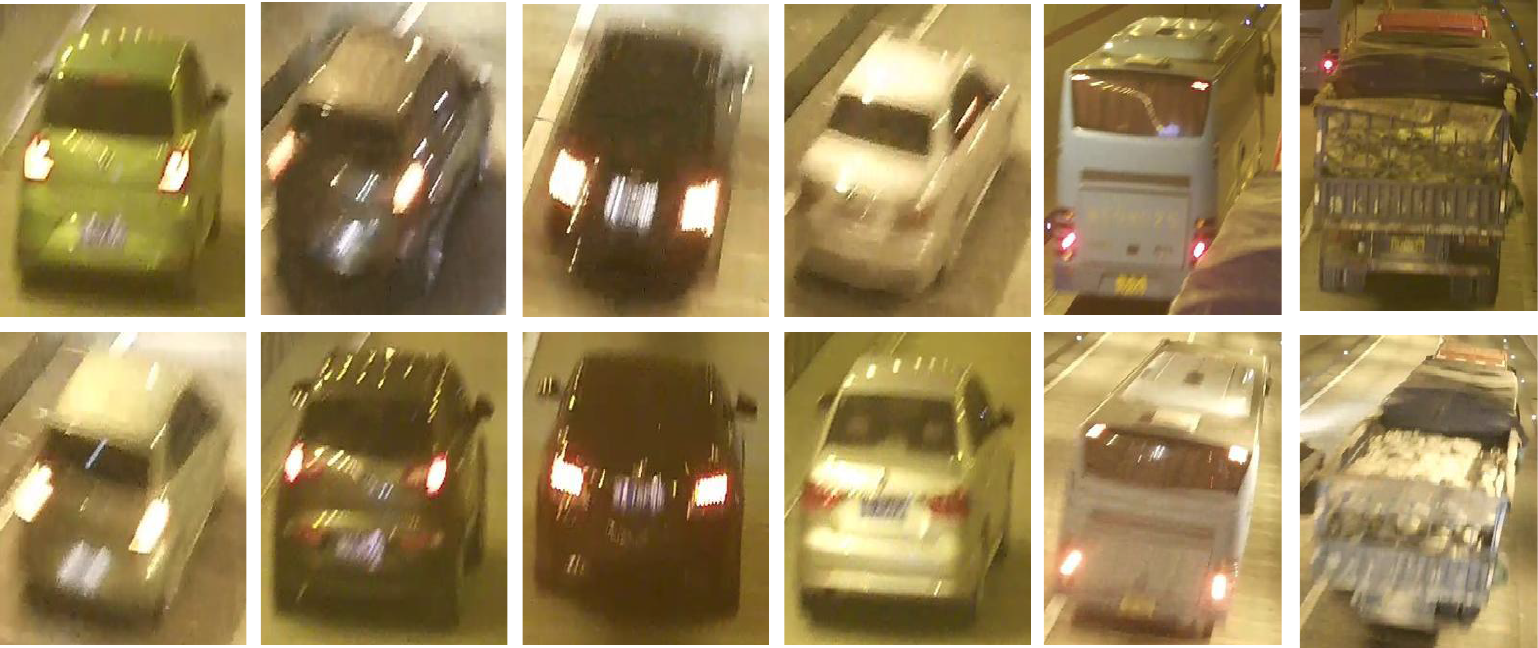}
  \caption{Some typical samples in the Tunnel-VReID dataset. Images in the same column represent the same vehicle.}
  \vspace{-1em}
  \label{fig5}
\end{figure}
In this work, we adopt three dataset for evaluation: the VeRi dataset \cite{LiuLMM2016}, the VehicleID dataset \cite{liuWPH2016} and the Tunnel-VReID dataset collected by ourselves. Some examples of Tunnel-VReID dataset are shown in Fig. \ref{fig5}, which show more frequent illumination and blurring challenges.

The VeRi dataset \cite{LiuLMM2016} contains over 50,000 images of 776 different vehicles captured by 20 cameras. Each vehicle is captured by 2$\sim$18 cameras in $1 km^2$ urban area. The dataset is split into 576 vehicles with 37,778 images for training and 200 vehicles with 11,579 images for testing. In the testing set, 1,678 images are selected as the query to retrieve corresponding images from the rest.

VehicleID dataset \cite{liuWPH2016} is a large-scale dataset collected during daytime in open road. It contains 221,567 images of 26,328 vehicles in total (8.42 images/vehicle in average), and is split into 13,134 vehicles for training and 13,133 vehicles for testing. Following \cite{liuWPH2016}, we use three test subsets of different sizes, \emph{i.e.}, small size with 7,332 images of 800 vehicles, medium size with 12,995 images of 1,600 vehicles, and large size with 20,038 images of 2,400 vehicles.

Tunnel-VReID is a new dataset that is collected from 9 pairs of 1920$\times$1080 HD surveillance cameras in three different expressway tunnels by ourselves. Tunnel-VReID dataset includes 1,000 pairs of vehicle identities, which can be used for the method evaluation for small scale dataset. Each vehicle identity contains two images captured by two non-overlapping cameras. Then we annotate bounding boxes for vehicles carefully. We take one camera view as the probe set, another one as the gallery set, and then randomly divide the pairs into equal half for training and testing. Tunnel-VReID dataset is available at: \url{https://github.com/ZHU912010/Tunnel-VReID-dataset}.

We adopt the mean average precision (mAP) and cumulative match curve (CMC) to evaluate the performance, where CMC represents the chance of correct match appearing in the top $1,2,. . . ,n$ of the ranked candidate list corresponding to Rank-$1,2,. . . ,n$, respectively. The mAP is the mean value of average precision of all queries reflecting both precision and recall of Re-ID.
\subsection{Implementation Details}
\subsubsection{Image-to-image style transfer}
Following CycleGAN \cite{ZhuPIE2017}, all images are resized to $256\times256$. The initial learning rates are $0.0002$ for generators and $0.0001$ for discriminators. For Tunnel-VReID dataset, the learning rates are linearly reduced to zero from $100$ epochs to $150$ epochs and from $20$ epochs to $30$ epochs for VeRi and VehicleID datasets with larger scale of data. $\alpha$ and $\beta$ in Eq.\ref{eq:2} are empirically set as $10$ and $5$, respectively.
\subsubsection{Baseline CNN model for Re-ID}
We build a triplet network as baseline CNN model for Re-ID. The batch size is set as 72, containing $K=12$ vehicle identities with $6$ multi-order images. In the training process, the learning rate is gradually decreased by exponentially decay after $t_0$ iterations, the schedule is:
\begin{equation}
\gamma(t)=\left\{\begin{array}{ll}{\gamma_{0}} & {\text { if } t \leq t_{0}} \\ {\gamma_{0}* 0.001^{\frac{t-t_{0}}{t_{1}-t_{0}}}} & {\text { if } t_{0} \leq t \leq t_{1}}\end{array}\right.,
\end{equation}
where $t$ denotes the number of iterations. Model stops training when reaching $t_1$. In this work, we set $t_0=20,000$ and $t_1=40,000$ for VeRi and VehicleID datasets, and $t_0=15,000$ and $t_1=20,000$ for Tunnel-VReID dataset. The Adam optimizer is adopted with the base learning rate $\gamma_0=0.0003$. During testing, we extract a 1024-dimensional vehicle embedding for each vehicle image and evaluate the performance.
Experiments are implemented based on the Pytorch platform on two NVIDIA GeForce GTX 2080 Ti with the GPU memory of 22GB.
\begin{table*}[htbp]
  \centering
  \scriptsize
  \caption{Rank-1 values ($\%$) of different cross-distance combinations on three vehicle Re-ID datasets. The best one is marked in bold fonts.}
  \renewcommand\arraystretch{2}
  \setlength{\tabcolsep}{0.7mm}{
    \begin{tabular}{c|ccccc|cccc|ccc|cc|c}
    \hline   \hline
    Rank-1 &$d_1$ &$d_2$ &$d_8$  &$d_{10}$ &$d_{11}$ &$d_1$+$d_2$  &$d_1$+$d_8$  &$d_1$+$d_{10}$ &$d_1$+$d_{11}$   & $d_1$+$d_2$+$d_8$ &$d_1$+$d_2$+$d_{10}$& $d_1$+$d_2$+$d_{11}$ & $d_1$+$d_2$+$d_8$+$d_{10}$ &$d_1$+$d_2$+$d_8$+$d_{11}$&$d_1$+$d_2$+$d_8$+$d_{10}$+$d_{11}$\\
    \hline
    VeRi         & 89.0 & 87.5 & 88.2  & 87.0 & 87.7& 89.7 & 90.2  & 89.6 & 89.6   & 89.6 & \textbf{91.4} &90.1& 89.8 & 89.5& 89.8 \\
    VehicleID    & 80.6 & 80.3  & 80.4  & 80.2 & 79.4& 81.6  & 81.4  & 81.5 & 80.7 & 81.6 & \textbf{82.5} &81.8 & 82.2& 82.1& 82.2\\
     Tunnel-VReID & 69.0  & 68.2    & 67.6    & 66.6  & 62.3& 79.6    & 77.9    & 78.3  & 77.5 & 79.1  & \textbf{80.5}  &80.0& 79.9& 78.9& 79.7 \\
    \hline   \hline
    \end{tabular}}
  \label{tab5}
\end{table*}
\subsection{Ablation Studies}\label{subsec:4-D}
\subsubsection{Effect of cross distance combination in testing}
Different cross distance combinations contain different information between domains. To find out the optimal combination of cross distance for query rank, we conduct a detailed comparative test on the three vehicle Re-ID datasets. Notably, this subsection aims to check which group of cross distances is the best in testing, where we re-trained the Re-ID model without the multi-order augmented center loss. Actually, as described in Section \ref{subsection:3}, there are $C_6^2-6=9$ kinds of cross distances that can be used for querying, which causes $\sum_{c=1}^9 C_9^{c}$ enumerations. However, for a query vehicle image $x$, in order to give a convenient and reasonable test, we treat the original $x$ as the basic element, and exploit the performance of other samples linking with it. Therefore, in our testing, $d_1, d_2$, and $d_8$ are firstly selected for testing. Specially, $d_{10}$ and $d_{11}$ indicate the relation within a common space approached from two different domains, and they may be helpful for Re-ID. Hence, we take them into the testing distance list for an attempt. The results are demonstrated in Table. \ref{tab5}. $d_1$ achieves the highest rank-1 score among five single cross distances, which indicates that the original images from different domains are essential. Considering more distances with $d_1$, the performance is boosted significantly until the combination $d_1+d_2+d_{10}$, and then decreases with more distances. Interestingly, the improvement margin on the Tunnel-VReID is larger then the one on the VeRi and VehicleID datasets. From Table. \ref{tab5}, we can see that our method is promising for the small-scale dataset. Additionally, since the $2^{nd}$-order samples may contain more noise than $1^{st}$-order after two times of generation, the performance shows degradation (fusing $d_8$ and $d_{11}$). Therefore, we select $d_1+d_2+d_{10}$ as the best combination.

\subsubsection{Role of multi-order augmentation}
In order to investigate the effectiveness of multi-order augmented images, we compared the proposed method after training with different order of images (also detached the center loss for a pure comparison). In the testing, we compared the Rank-1 value on three groups: $d_1$ with only $0$-order sample training; $d_1$ and $d_1+d_2+d_{10}$ with $0,1$-order sample training; $d_1$ and $d_1+d_2+d_{10}$  with $0,1,2$-order sample training. The results are listed in Table.  \ref{tab4}. From this table, the augmentation achieved significant performance gains, especially with the multi-order augmentation. The $1^{st}$-order images can only improve the accuracy of $d_1$ of $0^{th}$-order with an increase of $8.4\%$, $4.8\%$, and $2\%$ for Tunnel-VReID, VeRi and VehicleID, respectively, while $0^{th}$,$1^{st}$,$2^{nd}$-order images increased the performance gain with almost $27.1\%$, $15.4\%$ and $3.6\%$ for Tunnel-VReID, VeRi and VehicleID dataset. The performance on Tunnel-VReID with small scale samples is boosted prominently.
\begin{table}[htbp]
  \centering
  \scriptsize
  \renewcommand\arraystretch{1.5}
  \caption{Rank-1 values ($\%$) of different training data on three vehicle Re-ID datasets. The best one is highlighted in bold fonts.}
  \resizebox{82mm}{15mm}{
    \begin{tabular}{c|c|c|c|c}
    \hline\hline
    {Training Data}&Testing &VeRi & VehicleID & Tunnel-VReID  \\
    \hline
    \multirow{1}[0]{*}{$0^{th}$-order}   & $d_1$      &76.0 & 78.9     & 53.4 \\
    \hline
    \multirow{2}[0]{*}{$0^{th}$,$1^{st}$-order} & $d_1$      & 77.7 &79.5 & 58.0 \\

                              & $d_1+d_{2}+d_{10}$   & 80.8     & 80.9     &  61.8\\
    \hline
    \multirow{2}[0]{*}{$0^{th}$,$1^{st}$,$2^{nd}$-order} & $d_1$    &87.5   & 80.6 & 69.7  \\
                                & $d_1+d_{2}+d_{10}$ &   \textbf{91.4}   & \textbf{82.5}     & \textbf{80.5}  \\
    \hline\hline
    \end{tabular}}
    \vspace{-1em}
  \label{tab4}
\end{table}

\begin{table*}[htbp]\scriptsize
  \centering
  \caption{Performance (\%) of different methods on VeRi and VehicleID datasets. The best one is highlighted in bold fonts.}
    \begin{tabular}{c|c|c|c||c|c|c|c|c|c|c|c|c}
    \hline\hline
    \multicolumn{4}{c||}{VeRi}     & \multicolumn{7}{c}{VehicleID} \\
    \midrule
    \multirow{2}[4]{*}{Methods} & \multirow{2}[4]{*}{mAP} & \multirow{2}[4]{*}{Rank 1} & \multirow{2}[4]{*}{Rank5} & \multirow{2}[4]{*}{Methods} & \multicolumn{2}{c|}{Test size = 800} & \multicolumn{2}{c|}{Test size = 1600} & \multicolumn{2}{c|}{Test size = 2400} & \multicolumn{2}{c}{Average}\\
\cmidrule{6-13}          &       &       &       &       & Rank 1 & Rank 5 & Rank 1 & Rank 5 & Rank 1 & Rank 5&Rank 1 & Rank 5 \\
    \midrule
    LOMO\cite{LiaoHZ2015}  & 9.78  & 23.87 & 39.14 & LOMO\cite{LiaoHZ2015}  & 19.76 & 32.01 & 18.85 & 29.18 & 15.32 & 25.29 & 17.97 & 28.82\\
    DGD\cite{xiaoLOW2016}   & 17.92 & 50.7  & 67.52 & DGD\cite{xiaoLOW2016}   & 44.80 & 66.28 & 40.25 & 65.31 & 37.33 & 57.82 & 40.79 & 63.13\\
    FACT\cite{LiuLMM2016}  & 18.73 & 51.85 & 67.16 & FACT\cite{LiuLMM2016}  & 49.53 & 68.07 & 44.59 & 64.57 & 39.92 & 60.32 & 44.68 & 61.65\\
    GoogLeNet\cite{yangLCT2015} & 17.81 & 52.12 & 66.79 & GoogLeNet\cite{yangLCT2015} & 47.88 & 67.18 & 43.40 & 63.86 & 38.27 & 59.39 & 43.18 & 63.48\\
    XVGAN\cite{zhouYSL2017}  & 24.65 & 60.2  & 77.03 & VGG+CCL\cite{liuWPH2016} & 43.62 & 64.84 & 39.94 & 62.98 & 35.68 & 56.24 & 49.10 & 61.35\\
    OIFE\cite{WangTLYY2017}  & 48.00  & 65.92 & 87.66 & XVGAN\cite{zhouYSL2017}  & 52.87 & 80.83 & 49.55 & 71.39 & 44.89 & 66.65 & 44.89 & 72.96\\
    PROVID\cite{LiuLMM2018} & 53.42 & 81.56 & 95.11 & VAMI\cite{ZhouS2018}  & 63.12 & 83.25 & 52.87 & 75.12 & 47.34 & 70.29 & 57.44 & 76.22\\
    VAMI\cite{ZhouS2018}  & 50.13 & 77.03 & 90.82 & MLSR\cite{HouZCZCM2019}  & 65.78  & 78.09  & 64.24   & 73.11  & 60.05  & 70.81 & 63.35  & 74.00 \\
    PRND\cite{heLZT2019}  & 70.2  & 92.2  & \textbf{97.9}  &PRND\cite{heLZT2019}  & 78.4  & \textbf{92.3}  & 75.0   & 88.3  & 74.2  & 86.4  & 75.9  & 89.0\\
    \midrule
    Baseline & 51.2 & 76.0 & 86.4 & Baseline & 78.9 & 90.2 & 63.7 & 81.5 & 60.3 & 79.7& 67.6 & 83.8 \\
    Ours-T     & 69.7 & 91.4 & 95.7 & Ours-T  & 82.5 & 90.4 & 77.5 & 88.4 & 74.9 & 86.8 & 78.3 & 88.5\\
    Ours-T-C  & \textbf{70.4} & \textbf{92.8} & 96.8 & Ours-T-C  & \textbf{82.9} & 90.5 & \textbf{78.7} & \textbf{89.7} & \textbf{75.9} & \textbf{87.1} & \textbf{79.1} & \textbf{89.1}\\
    \hline\hline
    \end{tabular}
  \label{tab:addlabel}
\end{table*}
\subsubsection{Influence of center loss}
\begin{figure}
  \centering
  \includegraphics[width=\hsize]{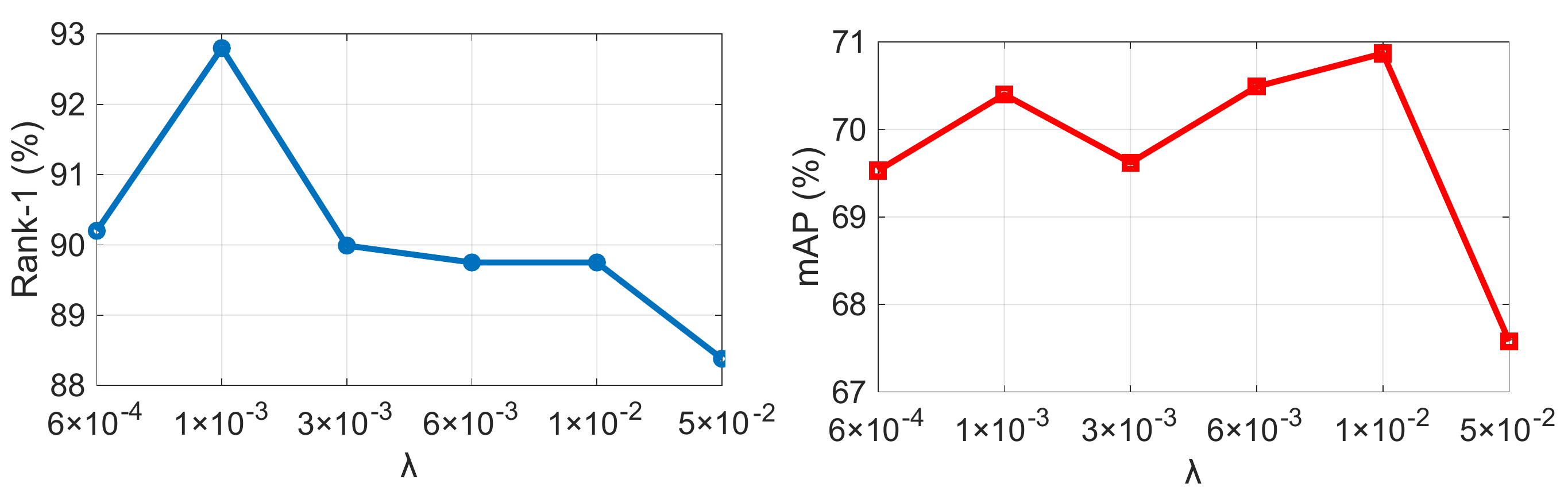}
  \caption{The Rank-1 and mAP values, w.r.t, $\lambda$ on VeRi dataset.}
    \vspace{-1.2em}
  \label{fig6}
\end{figure}

In order to check the influence of the center loss, we experimentally set the balancing parameter $\lambda$ as [0.0006, 0.001, 0.003, 0.006, 0.01]. The Performance of the proposed models on VeRi dataset are shown in Fig. ~\ref{fig6}. It is clear that proper value of $\lambda$ can improve the performance. We can observe that the Rank-1 value has the highest performance when $\lambda$=0.001, and the mAP has the second highest performance. After $\lambda$ is greater than 0.001, performance is no longer ascending and fluctuates. Therefore, we finally choose $\lambda$=0.001 as default setting.

\subsection{Comparison with State-of-the-art}
\subsubsection{Evaluation on VeRi and VehicleID Datasets}
We compare the proposed method (``Ours-T" without the center loss, and ``Our-T-C" with the center loss) with state-of-the-art vehicle re-ID methods on VeRi and VehicleID Datasets. They are DGD \cite{xiaoLOW2016}, GoogLeNet \cite{yangLCT2015}, FACT \cite{LiuLMM2016}, VGG+CCL\cite{liuWPH2016} (VehicleID dataset only), PROVID \cite{LiuLMM2018} (VeRi dataset only), OIFE \cite{WangTLYY2017} (VeRi dataset only), XVGAN \cite{zhouYSL2017}, VAMI \cite{ZhouS2018}, MSLR \cite{HouZCZCM2019} (VehicleID dataset only) and the newest PRND \cite{heLZT2019}.

\begin{figure*}
  \centering
  \includegraphics[width=0.93\hsize]{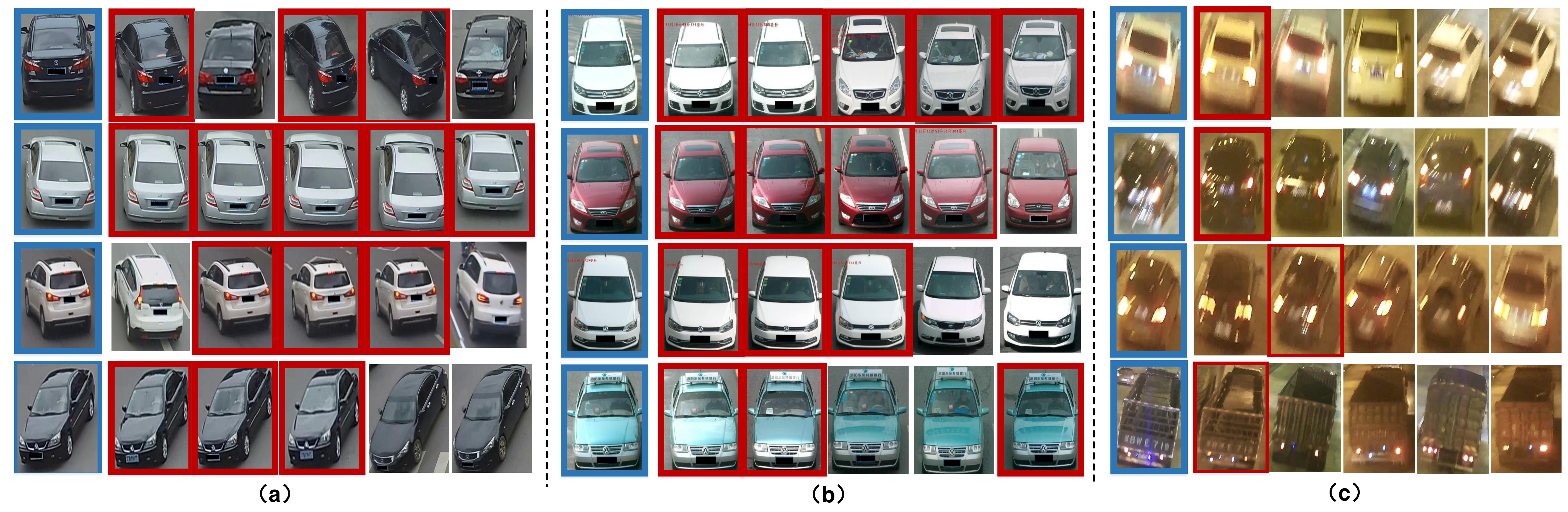}
  \caption{Examples of Re-ID retrieval results by the proposed method on the (a) VeRi, (b) VehicleID, and (c) Tunnel-VReID datasets. For each dataset, each row represents the ranking results with the first image being the query and the rest ones being the returned list. The query images and its correct match images are marked by blue and red rectangle boxes, respectively.}
  \vspace{-1em}
  \label{fig7}
\end{figure*}

As listed in Table~\ref{tab:addlabel}, the results show that the proposed method (Ours) achieves the highest mAP of $70.4\%$ and Rank-1 accuracy of $92.8\%$, as well as the second Rank-5 accuracy of $96.8\%$ on VeRi dataset, and obtain the highest performance on small-, middle- and large-scale versions of VehicleID dataset. LOMO \cite{LiaoHZ2015} based on hand-crafted features is weaker than deep learning based methods because of huge variations in large scale dataset. Compared with those deep learning methods, our baseline obtains a competitive performance, where most of the best performance are generated by our full model (Ours+CenterLoss). The newest PRND \cite{heLZT2019} owns a similar ability for Re-ID performance with ours, while its testing stage needs to pre-detect the semantic parts of the vehicle while our method only needs to make a global cross-distance comparison. Fig. \ref{fig7} shows some snapshots of Re-ID results, where almost top three positives are correctly queried by our method.
\subsubsection{Evaluation on Tunnel-VReID Dataset}
\begin{table}[htbp]
\scriptsize
\renewcommand{\arraystretch}{1.5}
\caption{Performance (\%) of different methods on our Tunnel-VReID dataset. The best one is highlighted in bold fonts.}
\label{tab1}
\centering
\setlength{\tabcolsep}{2mm}{
\begin{tabular}{c|c|c|c}
\hline\hline
Method   & Rank-1 & Rank-5 & Rank-10 \\
\hline
XQDA\cite{LiaoHZ2015}           & 40.0 & 71.1 & 81.5 \\
MLAPG\cite{LiaoL2015}           & 41.6 & 71.9 & 83.1 \\
VGG+Triplet Loss\cite{Ding2015} & 43.6 & 69.6 & 77.8 \\
CNN Embedding\cite{zheng2018}   & 45.8 & 71.1 & 79.2 \\
PCB\cite{SunZYTW2018}           & 48.9 & 73.2 & 80.8 \\
PRND\cite{heLZT2019}            & 68.2  & 84.2  & 89.3\\
\hline
Baseline                        & 53.4 & 78.2 & 87.2 \\
Ours-T                            & 80.5 & 87.9 & 91.4 \\
Ours-T-C                & \textbf{81.3} & \textbf{88.5} & \textbf{92.0} \\
\hline\hline
\end{tabular}}
\vspace{-1em}
\end{table}
Beside the VeRi and the VehicleID datasets collected from daytime and open scene, we further evaluate our method by the Tunnel-VReID dataset collected by ourselves on tunnel scene, where dim illumination and motion blur are frequent on this kind of scene. LOMO \cite{LiaoHZ2015}, MLAPG \cite{LiaoL2015}, VGG+Triplet Loss \cite{Ding2015}, CNN Embedding \cite{zheng2018}, PCB \cite{SunZYTW2018} and PRND \cite{heLZT2019} are selected as the competition list. Table~\ref{tab1} shows the rank-1, 5, 10 value comparisons. From the results, we can observe that the proposed method outperforms the other ones in the rank-1, 5, 10 scores. Our method with multi-order data argumentation shows a performance gain of $13.1\%$ on the newest PRND \cite{heLZT2019}. That is because that Tunnel-VReID dataset with only $1000$ pair of images is a small-scale dataset. The deep learning based methods without argumentation may be limited for this situation, where our baseline without argumentation degrades significantly with near $28\%$ margin, which further proves the necessity of multi-order deep cross-distance learning. Especially for the blurred factors, our model can achieve a promising discrimination, as shown in Fig. \ref{fig7}.

\section{Conclusions}\label{section5}
In this paper, we formulated a multi-order deep cross-distance learning (DCDLearn) model for vehicle re-identification, which exploited the cross-relation of multi-order images consisting of the real, style-transferred and reconstructed images within the circle of our designed one-view CycleGAN. One-view CycleGAN is efficient and can avoid the enumerative and overhead pair-wise CycleGANs in previous Re-ID works. Through DCDLearn, optimal vehicle representations of multi-order images were learned and reduced the cross-camera discrepancy for vehicle Re-ID. Qualitative and quantitative experiments demonstrated the effectiveness of our method. In the future, we plan to extend this work to cross-dataset Re-ID task.

{\small{
\bibliographystyle{ieee_fullname}
\bibliography{bibfile}}

\begin{thebibliography}{10}\itemsep=-1pt

\bibitem{Christos2008}
Christos{-}Nikolaos Anagnostopoulos, Ioannis Anagnostopoulos, I.~D. Psoroulas,
  Vassilis Loumos, and Eleftherios Kayafas.
\newblock License plate recognition from still images and video sequences: {A}
  survey.
\newblock {\em {IEEE} Trans. Intelligent Transportation Systems},
  9(3):377--391, 2008.

\bibitem{BaiLGWWD2018}
Yan Bai, Yihang Lou, Feng Gao, Shiqi Wang, Yuwei Wu, and Ling-Yu Duan.
\newblock Group-sensitive triplet embedding for vehicle reidentification.
\newblock {\em {IEEE} Trans. Multimedia}, 20(9):2385--2399, 2018.

\bibitem{bousmalis2017}
Konstantinos Bousmalis, Nathan Silberman, David Dohan, Dumitru Erhan, and Dilip
  Krishnan.
\newblock Unsupervised pixel-level domain adaptation with generative
  adversarial networks.
\newblock In {\em Proc. IEEE Conference on Computer Vision and Pattern
  Recognition}, pages 3722--3731, 2017.

\bibitem{ChoiCKHKC2018}
Yunjey Choi, Minje Choi, Munyoung Kim, Jung-Woo Ha, Sunghun Kim, and Jaegul
  Choo.
\newblock {StarGAN}: Unified generative adversarial networks for multi-domain
  image-to-image translation.
\newblock In {\em Proc. IEEE Conference on Computer Vision and Pattern
  Recognition}, pages 8789--8797, 2018.

\bibitem{chungD2019}
Dahjung Chung and Edward~J Delp.
\newblock Camera-aware image-to-image translation using similarity preserving
  stargan for person re-identification.
\newblock In {\em Proc. IEEE Conference on Computer Vision and Pattern
  Recognition Workshops}, pages 0--0, 2019.

\bibitem{DengZYKYJ2018}
Weijian Deng, Liang Zheng, Qixiang Ye, Guoliang Kang, Yi Yang, and Jianbin
  Jiao.
\newblock Image-image domain adaptation with preserved self-similarity and
  domain-dissimilarity for person re-identification.
\newblock In {\em Proc. IEEE Conference on Computer Vision and Pattern
  Recognition}, pages 994--1003, 2018.

\bibitem{Ding2015}
Shengyong Ding, Liang Lin, Guangrun Wang, and Hongyang Chao.
\newblock Deep feature learning with relative distance comparison for person
  re-identification.
\newblock {\em Pattern Recognition}, 48(10):2993--3003, 2015.

\bibitem{goodfellow2014}
Ian Goodfellow, Jean Pouget-Abadie, Mehdi Mirza, Bing Xu, David Warde-Farley,
  Sherjil Ozair, Aaron Courville, and Yoshua Bengio.
\newblock Generative adversarial nets.
\newblock In {\em Advances in Neural Information Processing Systems}, pages
  2672--2680, 2014.

\bibitem{GouWYL2016}
Chao Gou, Kunfeng Wang, Yanjie Yao, and Zhengxi Li.
\newblock Vehicle license plate recognition based on extremal regions and
  restricted boltzmann machines.
\newblock {\em {IEEE} Trans. Intelligent Transportation Systems},
  17(4):1096--1107, 2016.

\bibitem{heLZT2019}
Bing He, Jia Li, Yifan Zhao, and Yonghong Tian.
\newblock Part-regularized near-duplicate vehicle re-identification.
\newblock In {\em Proceedings of the IEEE Conference on Computer Vision and
  Pattern Recognition}, pages 3997--4005, 2019.

\bibitem{heKaiming2016}
Kaiming He, Xiangyu Zhang, Shaoqing Ren, and Jian Sun.
\newblock Deep residual learning for image recognition.
\newblock In {\em Proc. IEEE Conference on Computer Vision and Pattern
  Recognition}, pages 770--778, 2016.

\bibitem{HouZCZCM2019}
Jinhui Hou, Huanqiang Zeng, Lei Cai, Jianqing Zhu, Jing Chen, and Kai{-}Kuang
  Ma.
\newblock Multi-label learning with multi-label smoothing regularization for
  vehicle re-identification.
\newblock {\em Neurocomputing}, 345:15--22, 2019.

\bibitem{isola2017}
Phillip Isola, Jun-Yan Zhu, Tinghui Zhou, and Alexei~A Efros.
\newblock Image-to-image translation with conditional adversarial networks.
\newblock In {\em Proc. IEEE Conference on Computer Vision and Pattern
  Recognition}, pages 1125--1134, 2017.

\bibitem{kim2017}
Taeksoo Kim, Moonsu Cha, Hyunsoo Kim, Jung~Kwon Lee, and Jiwon Kim.
\newblock Learning to discover cross-domain relations with generative
  adversarial networks.
\newblock In {\em Proc. International Conference on Machine Learning}, pages
  1857--1865, 2017.

\bibitem{ledig2017}
Christian Ledig et~al.
\newblock Photo-realistic single image super-resolution using a generative
  adversarial network.
\newblock In {\em Proc. IEEE Conference on Computer Vision and Pattern
  Recognition}, pages 4681--4690, 2017.

\bibitem{liW2016}
Chuan Li and Michael Wand.
\newblock Precomputed real-time texture synthesis with markovian generative
  adversarial networks.
\newblock In {\em Proc. European Conference on Computer Vision}, pages
  702--716, 2016.

\bibitem{LiYJL2017}
Xiying Li, Minxian Yuan, Qianyin Jiang, and Guoming Li.
\newblock {VRID-1:} {A} basic vehicle re-identification dataset for similar
  vehicles.
\newblock In {\em Proc. IEEE International Conference on Intelligent
  Transportation Systems}, pages 1--8, 2017.

\bibitem{LiaoHZ2015}
Shengcai Liao, Yang Hu, Xiangyu Zhu, and Stan~Z. Li.
\newblock Person re-identification by local maximal occurrence representation
  and metric learning.
\newblock In {\em Proc. IEEE Conference on Computer Vision and Pattern
  Recognition}, pages 2197--2206, 2015.

\bibitem{LiaoL2015}
Shengcai Liao and Stan~Z. Li.
\newblock Efficient {PSD} constrained asymmetric metric learning for person
  re-identification.
\newblock In {\em Proc. IEEE International Conference on Computer Vision},
  pages 3685--3693, 2015.

\bibitem{liuWPH2016}
Hongye Liu, Yonghong Tian, Yaowei Wang, Lu Pang, and Tiejun Huang.
\newblock Deep relative distance learning: Tell the difference between similar
  vehicles.
\newblock In {\em Proc. IEEE Conference on Computer Vision and Pattern
  Recognition}, pages 2167--2175, 2016.

\bibitem{LiuZCHW2019}
Jiawei Liu, Zheng-Jun Zha, Di Chen, Richang Hong, and Meng Wang.
\newblock Adaptive transfer network for cross-domain person re-identification.
\newblock In {\em Proc. IEEE Conference on Computer Vision and Pattern
  Recognition}, pages 7202--7211, 2019.

\bibitem{LiuLMF2016}
Xinchen Liu, Wu Liu, Huadong Ma, and Huiyuan Fu.
\newblock Large-scale vehicle re-identification in urban surveillance videos.
\newblock In {\em Proc. IEEE International Conference on Multimedia and Expo},
  pages 1--6, 2016.

\bibitem{LiuLMM2018}
Xinchen Liu, Wu Liu, and Tao Mei.
\newblock {PROVID}: Progressive and multimodal vehicle reidentification for
  large-scale urban surveillance.
\newblock {\em {IEEE} Trans. Multimedia}, 20(3):645--658, 2018.

\bibitem{LiuLMM2016}
Xinchen Liu, Wu Liu, Tao Mei, and Huadong Ma.
\newblock A deep learning-based approach to progressive vehicle
  re-identification for urban surveillance.
\newblock In {\em Proc. European Conference on Computer Vision}, pages
  869--884, 2016.

\bibitem{PengWZF2019}
Jinjia Peng, Huibing Wang, Tongtong Zhao, and Xianping Fu.
\newblock Learning multi-region features for vehicle re-identification with
  context-based ranking method.
\newblock {\em Neurocomputing, DOI: 10.1016/j.neucom.2019.06.013}, 2019.

\bibitem{shanSK2008}
Ying Shan, Harpreet~S Sawhney, and Rakesh Kumar.
\newblock Unsupervised learning of discriminative edge measures for vehicle
  matching between nonoverlapping cameras.
\newblock {\em {IEEE} Trans. Pattern Analysis and Machine Intelligence},
  30(4):700--711, 2008.

\bibitem{ShenXLYW2017}
Yantao Shen, Tong Xiao, Hongsheng Li, Shuai Yi, and Xiaogang Wang.
\newblock Learning deep neural networks for vehicle re-id with
  visual-spatio-temporal path proposals.
\newblock In {\em Proc. IEEE International Conference on Computer Vision},
  pages 1918--1927, 2017.

\bibitem{SunZYTW2018}
Yifan Sun, Liang Zheng, Yi Yang, Qi Tian, and Shengjin Wang.
\newblock Beyond part models: Person retrieval with refined part pooling (and a
  strong convolutional baseline).
\newblock In {\em Proc. European Conference on Computer Vision}, pages
  501--518, 2018.

\bibitem{TaigmanPW2017}
Yaniv Taigman, Adam Polyak, and Lior Wolf.
\newblock Unsupervised cross-domain image generation.
\newblock In {\em Proc. International Conference on Learning Representations},
  2017.

\bibitem{tangZL2019}
Haotian Tang, Yiru Zhao, and Hongtao Lu.
\newblock Unsupervised person re-identification with iterative self-supervised
  domain adaptation.
\newblock In {\em Proc. IEEE Conference on Computer Vision and Pattern
  Recognition Workshops}, pages 0--0, 2019.

\bibitem{TangWJZL2017}
Yi Tang, Di Wu, Zhi Jin, Wenbin Zou, and Xia Li.
\newblock Multi-modal metric learning for vehicle re-identification in traffic
  surveillance environment.
\newblock In {\em Proc. IEEE International Conference on Image Processing},
  pages 2254--2258, 2017.

\bibitem{WangTLYY2017}
Zhongdao Wang, Luming Tang, Xihui Liu, Zhuliang Yao, Shuai Yi, Jing Shao,
  Junjie Yan, Shengjin Wang, Hongsheng Li, and Xiaogang Wang.
\newblock Orientation invariant feature embedding and spatial temporal
  regularization for vehicle re-identification.
\newblock In {\em Proc. IEEE International Conference on Computer Vision},
  pages 379--387, 2017.

\bibitem{WangYYBS2018}
Zheng Wang, Mang Ye, Fan Yang, Xiang Bai, and Shin'ichi Satoh.
\newblock Cascaded {SR-GAN} for scale-adaptive low resolution person
  re-identification.
\newblock In {\em Proc. International Joint Conference on Artificial
  Intelligence}, pages 3891--3897, 2018.

\bibitem{WeiZGT2018}
Longhui Wei, Shiliang Zhang, Wen Gao, and Qi Tian.
\newblock Person transfer gan to bridge domain gap for person
  re-identification.
\newblock In {\em Proc. IEEE Conference on Computer Vision and Pattern
  Recognition}, pages 79--88, 2018.

\bibitem{WenLYZDS2011}
Ying Wen, Yue Lu, Jingqi Yan, Zhenyu Zhou, Karen~M. von Deneen, and Pengfei
  Shi.
\newblock An algorithm for license plate recognition applied to intelligent
  transportation system.
\newblock {\em {IEEE} Trans. Intelligent Transportation Systems},
  12(3):830--845, 2011.

\bibitem{wen2016discriminative}
Yandong Wen, Kaipeng Zhang, Zhifeng Li, and Yu Qiao.
\newblock A discriminative feature learning approach for deep face recognition.
\newblock In {\em Proc. European Conference on Computer Vision}, pages
  499--515, 2016.

\bibitem{xiaoLOW2016}
Tong Xiao, Hongsheng Li, Wanli Ouyang, and Xiaogang Wang.
\newblock Learning deep feature representations with domain guided dropout for
  person re-identification.
\newblock In {\em Proc. IEEE Conference on Computer Vision and Pattern
  Recognition}, pages 1249--1258, 2016.

\bibitem{yanTWZH2017}
Ke Yan, Yonghong Tian, Yaowei Wang, Wei Zeng, and Tiejun Huang.
\newblock Exploiting multi-grain ranking constraints for precisely searching
  visually-similar vehicles.
\newblock In {\em Proc. IEEE International Conference on Computer Vision},
  pages 562--570, 2017.

\bibitem{yangLCT2015}
Linjie Yang, Ping Luo, Chen Change~Loy, and Xiaoou Tang.
\newblock A large-scale car dataset for fine-grained categorization and
  verification.
\newblock In {\em Proc. IEEE Conference on Computer Vision and Pattern
  Recognition}, pages 3973--3981, 2015.

\bibitem{yiZTG2017}
Zili Yi, Hao Zhang, Ping Tan, and Minglun Gong.
\newblock Dualgan: Unsupervised dual learning for image-to-image translation.
\newblock In {\em Proc. IEEE International Conference on Computer Vision},
  pages 2849--2857, 2017.

\bibitem{zapletalH2016}
Dominik Zapletal and Adam Herout.
\newblock Vehicle re-identification for automatic video traffic surveillance.
\newblock In {\em Proc. IEEE Conference on Computer Vision and Pattern
  Recognition Workshops}, pages 25--31, 2016.

\bibitem{zhaoSWC2019}
Yanzhu Zhao, Chunhua Shen, Huibing Wang, and Shengyong Chen.
\newblock Structural analysis of attributes for vehicle re-identification and
  retrieval.
\newblock {\em IEEE Transactions on Intelligent Transportation Systems, DOI:
  10.1109/TITS.2019.2896273}, 2019.

\bibitem{zheng2018}
Zhedong Zheng, Liang Zheng, and Yi Yang.
\newblock A discriminatively learned {CNN} embedding for person
  re-identification.
\newblock {\em {ACM} Trans. Multimedia Computing, Communications, and
  Applications (TOMM)}, 14(1):13:1--13:20, 2018.

\bibitem{ZhongZZLY2018}
Zhun Zhong, Liang Zheng, Zhedong Zheng, Shaozi Li, and Yi Yang.
\newblock Camera style adaptation for person re-identification.
\newblock In {\em Proc. IEEE Conference on Computer Vision and Pattern
  Recognition}, pages 5157--5166, 2018.

\bibitem{ZhongZZLY2019}
Zhun Zhong, Liang Zheng, Zhedong Zheng, Shaozi Li, and Yi Yang.
\newblock Cam{S}tyle: {A} novel data augmentation method for person
  re-identification.
\newblock {\em {IEEE} Trans. Image Processing}, 28(3):1176--1190, 2019.

\bibitem{ZhouLS2018}
Yi Zhou, Li Liu, and Ling Shao.
\newblock Vehicle re-identification by deep hidden multi-view inference.
\newblock {\em {IEEE} Trans. Image Processing}, 27(7):3275--3287, 2018.

\bibitem{zhouYSL2017}
Yi Zhou and Ling Shao.
\newblock Cross-view gan based vehicle generation for re-identification.
\newblock In {\em Proc. British Machine Vision Conference}, volume~1, pages
  1--12, 2017.

\bibitem{ZhouS2018}
Yi Zhou and Ling Shao.
\newblock Viewpoint-aware attentive multi-view inference for vehicle
  re-identification.
\newblock In {\em Proc. IEEE Conference on Computer Vision and Pattern
  Recognition}, pages 6489--6498, 2018.

\bibitem{zhuZHLLCZ2019}
Jianqing Zhu, Huanqiang Zeng, Jingchang Huang, Shengcai Liao, Zhen Lei, Canhui
  Cai, and Lixin Zheng.
\newblock Vehicle re-identification using quadruple directional deep learning
  features.
\newblock {\em {IEEE} Trans. Intelligent Transportation Systems, DOI:
  10.1109/TITS.2019.2901312}, 2019.

\bibitem{ZhuPIE2017}
Jun-Yan Zhu, Taesung Park, Phillip Isola, and Alexei~A Efros.
\newblock Unpaired image-to-image translation using cycle-consistent
  adversarial networks.
\newblock In {\em Proc. IEEE International Conference on Computer Vision},
  pages 2223--2232, 2017.

\end{thebibliography}
}

\end{document}